%% file: iclr2022_conference.tex
\documentclass{article} % For LaTeX2e
\usepackage{aloe2022,times}

\input{math_commands.tex}

\usepackage{hyperref}
\usepackage{url}
\usepackage{graphicx}
\usepackage{float}
\usepackage{multirow}
\usepackage{pifont}
\usepackage{enumitem}
\usepackage{algorithm}
\usepackage{algpseudocode}
\usepackage{bm}
\usepackage{wrapfig}
\usepackage{subcaption}

\title{Neuroevolution of Recurrent Architectures on Control Tasks}

\vspace{-1em}

\author{Maximilien Le Clei \& Pierre Bellec\\
Montreal University Geriatric Institute\\
\texttt{\{maximilien.le.clei,pierre.bellec\}@umontreal.ca}}

\iclrfinalcopy % Uncomment for camera-ready version, but NOT for submission.
\begin{document}

\maketitle

\vspace{-1.3em}

\begin{abstract}
\vspace{-1em}
Modern artificial intelligence works typically train the parameters of fixed-sized deep neural networks using gradient-based optimization techniques. Simple evolutionary algorithms have recently been shown to also be capable of optimizing deep neural network parameters, at times matching the performance of gradient-based techniques, e.g. in reinforcement learning settings. In addition to optimizing network parameters, many evolutionary computation techniques are also capable of progressively constructing network architectures. However, constructing network architectures from elementary evolution rules has not yet been shown to scale to modern reinforcement learning benchmarks. In this paper we therefore propose a new approach in which the architectures of recurrent neural networks dynamically evolve according to a small set of mutation rules. We implement a massively parallel evolutionary algorithm and run experiments on all 19 \mbox{OpenAI} Gym state-based reinforcement learning control tasks. We find that in most cases, dynamic agents match or exceed the performance of gradient-based agents while utilizing orders of magnitude fewer parameters. We believe our work to open avenues for real-life applications where network compactness and autonomous design are of critical importance. We provide our source code, final model checkpoints and full results at \textcolor{blue}{\href{http://www.github.com/MaximilienLC/nra/}{github.com/MaximilienLC/nra/}}.
\end{abstract}

\vspace{-1.5em}
\section{Introduction}
\vspace{-1em}
Artificial neural networks are computing systems that have become central to the field of artificial intelligence. Through waves of innovation, these networks have gotten bigger, more efficient, and increasingly competent on many tasks such as image classification \citep{dai2021coatnet}, image generation \citep{ramesh2021zero} and language modelling \citep{rae2021scaling}. Most often, the parameters of these artificial neural networks are optimized using first-order gradient-based optimization techniques, yet their architecture is for the most part still constructed manually.

In contrast, many evolutionary algorithm formulations make it possible to not only optimize a neural network's parameters but also construct its architecture \citep{stanley2019designing}. And while neuroevolution methods have been overshadowed by gradient-based techniques over the past decade, recent works have shown that evolutionary optimization is competitive with gradient-based learning at optimizing deep neural networks on various reinforcement learning problems \citep{salimans2017evolution,such2017deep,risi2019deep}. These advances were notably achieved by stripping down many mechanisms popular in traditional evolutionary methods, like agent crossover and speciation, to instead focus on leveraging larger-scale computational resources. 

These recent works however only focused on optimizing the neural network parameters, and did not attempt to evolve the network architectures. In this work, we therefore propose to investigate whether evolving neural network architectures can also lead to high performance on popular reinforcement learning tasks. We present a dynamic architecture framework for recurrent neural networks, embedded in a simple evolutionary algorithm, in which agents sample one of four structural mutations every iteration, enabling a population of agents to both increase and decrease their network capacity as the optimization process unfolds.

In order to evaluate these networks of dynamic capacity, we propose to run reinforcement learning experiments on all 19 \mbox{OpenAI} Gym \citep{brockman2016openai} state-based control tasks. We first compare the dynamic networks with standard static-sized deep neural networks optimized through the same evolutionary algorithm and find the dynamic networks generally more efficient and performant than the static networks. And while the static networks that we evolve are already relatively small by modern standards, averaging at $\sim$8,000 parameters, we find the final elite dynamic networks to be of even lower capacity, averaging at 127 parameters across all tasks. Interestingly, we also find the dynamic networks to be competitive with gradient-based techniques on the majority of these tasks.
\vspace{-1em}
\section{Related Work}
\vspace{-1em}
Reinforcement learning \citep{sutton2018reinforcement} is an artificial intelligence paradigm where artificial agents learn to maximize some notion of cumulative reward in an environment. Over the past decade, two traditional classes of techniques, Q-learning \citep{watkins1992q} and policy gradient methods \citep{sutton2000policy}, have been most successful in leveraging the representational power of deep neural networks, resulting in some of the most capable reinforcement learning agents to date \citep{mnih2013playing, lillicrap2015continuous, mnih2016asynchronous, schulman2017proximal, dabney2018distributional, haarnoja2018soft, fujimoto2018addressing, wang2020truly, kuznetsov2020controlling}.  

In recent years however, various works have shown many settings where, given sufficient computational resources, relatively simple evolutionary algorithms become competitive with gradient-based techniques at optimizing deep reinforcement learning agents. \cite{salimans2017evolution} first made use of an evolution strategy algorithm in order to optimize deep neural network parameters, resulting in competitive agents on both Atari and MuJoCo tasks. \cite{such2017deep} then showed that a simple genetic algorithm, stripped down of popular traditional mechanisms like crossovers and speciation, could also evolve deep neural network parameters in order to play many Atari games and control a humanoid on MuJoCo. Finally, \cite{risi2019deep} made use of a slightly more complex genetic algorithm in order to optimize the parameters of a World Model \citep{ha2018world}, in doing so evolving a competitive virtual car racing agent. 

In this work however, we solely focus on feature-based discrete and continuous control tasks, a setting explored many times over in the field evolutionary robotics \citep{doncieux2015evolutionary}. More precisely, we take part in the smaller subset of algorithms evolving both network architecture and parameters. While evolving both of these components has thoroughly been investigated before the turn of the century \citep{yao1999evolving}, this field is nowadays best known through the NEAT algorithm \citep{stanley2002evolving} for being most capable at solving a multitude of traditional reinforcement learning problems, e.g. the double pole balancing task \citep{stanley2004efficient}. Since then, NEAT has been extended to HyperNEAT \citep{stanley2009hypercube} in order to evolve larger scale neural networks that have been quite successfully applied to various robotics tasks such as evolving the gaits of both digital \citep{clune2009evolving} and real-life \citep{yosinski2011evolving} quadrupeds, and more recently, extended to evolve a 6-legged digital robot \citep{huizinga2016does}.

Our work is most similar to the NEAT algorithm, but makes the following novel contributions. Our algorithm is first of all less complex, not utilizing speciation and crossover components, and moreover very scalable, conveniently able to leverage large computational resources. The approach is also more general, utilizing fewer hyperparameters, and more flexible, allowing networks to not only increase but also decrease in capacity throughout the optimization process.

\vspace{-1em}
\section{Dynamic Recurrent Architectures} \label{methods}
\vspace{-1em}

In this section, we present recurrent neural networks whose architectures dynamically vary in capacity throughout the evolutionary process. Any such network is structured as a directed layered graph and is composed of a variable number of nodes and connections. In the first layer of any dynamic network, input nodes have the potential to transmit input information towards any non-input node located further in the graph. In the last layer, output nodes can potentially receive information from any existing node. Additionally, output nodes get to emit information out of the system and possibly towards any non-input node. Finally, in between these two layers ought to grow hidden nodes that can both receive information from any node and emit towards any non-input node.

In these networks, information flows through forward passes during which, layer after layer, non-input nodes make use of their mutable parameters (weight vector $w$ and bias $b$) in order to perform linear transformations of their input, followed by non-linear rectifier activations\footnote{$f(x) = \text{max}(0, w^{\top}x+b)$}. If a connection between two nodes is headed forward in the graph (meaning towards the output layer), the receiving node gets to make use of the emitting node's latest output information during the same network pass. However, if a connection between nodes is headed backward or to the same layer, the receiving node only gets to use the emitting node's latest output information during the next network pass.

Upon initialization, networks are always set to create the minimal amount of structure to fit the task at hand: a layer of input nodes and a layer of output nodes, both void of connections. In order to model more complex functions, these networks evolve in complexity through four structural mutations:

\;\;\;1) Grow Connection (\textcolor{violet}{Section \ref{grow_connection}})

\vspace{-.5em}\;\;\;2) Prune Connection (\textcolor{violet}{Section \ref{prune_connection}})

\vspace{-.5em}\;\;\;3) Grow Node (\textcolor{violet}{Section \ref{grow_node}})

\vspace{-.5em}\;\;\;4) Prune Node (\textcolor{violet}{Section \ref{prune_node}})

Before describing these four mutations, we define the following:\\
\textbf{In-nodes}: \textit{($\neq$ input nodes)} Set of nodes that a given node receives information from. Each node possesses its unique set of in-nodes.\\
\textbf{Out-nodes}: \textit{($\neq$ output nodes)} Set of nodes that a given node emits information to. Each node possesses its unique set of out-nodes.\\
\textbf{Receiving nodes}: Set of all input nodes plus all hidden/output nodes that possess in-nodes.\\
\textbf{Emitting nodes}: List of nodes possessing out-nodes. Nodes appear in this list once per out-node that they possess.

\vspace{.1em} % might want to get rid of later
\begin{minipage}{\linewidth}
\centering
\begin{minipage}{0.484\linewidth}

\subsection{Mutation \#1 : Grow Connection} \label{grow_connection}
\textbf{Step 1.} A first node is sampled\footnotemark\;from the set of all receiving nodes.\\
\textbf{Step 2.} A second node is sampled from the set of all hidden and output nodes.\\
\textbf{Step 3.} A new weighted\footnotemark\;connection is formed from the first to the second sampled node.\\\\
We give an example of this mutation in \textcolor{violet}{Figure \ref{fig:grow_connection}}.
\begin{figure}[H]
\includegraphics[width=.98\linewidth]{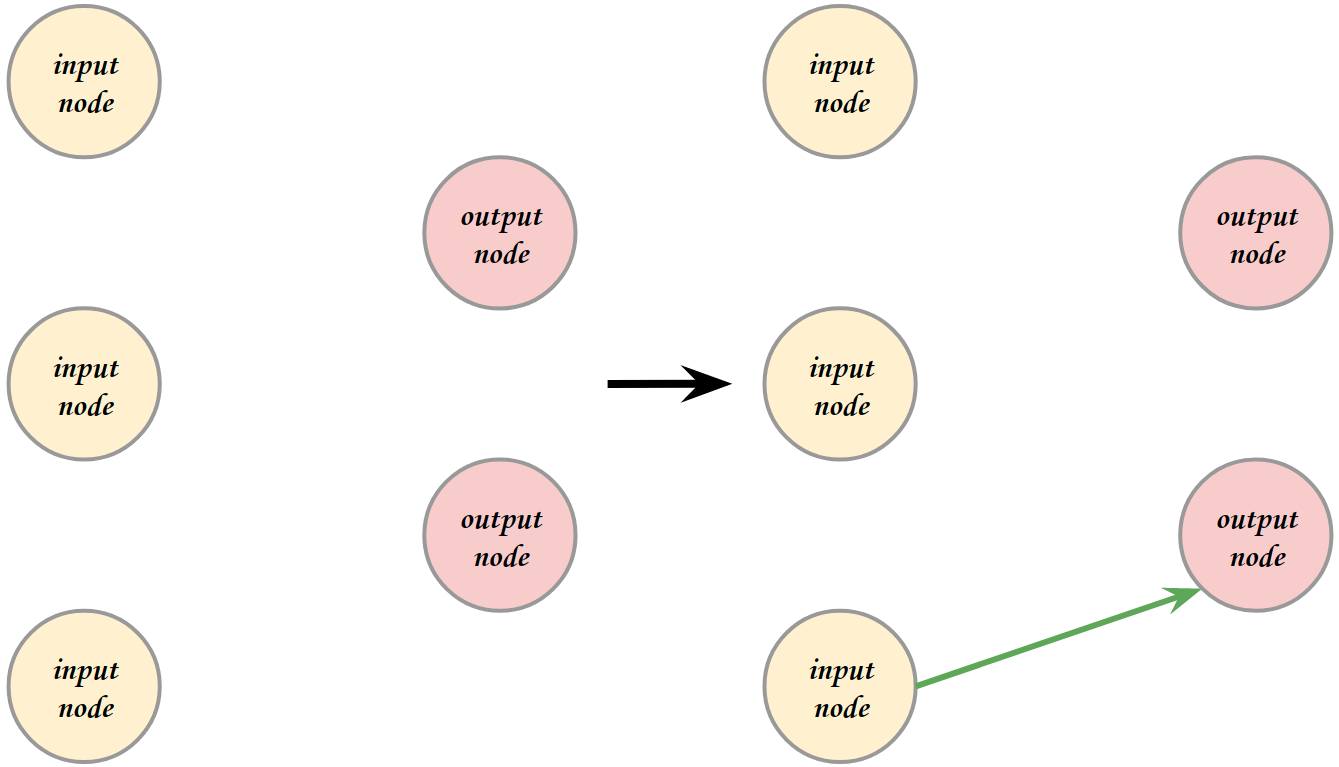}
\caption{\textbf{Mutation \#1 : Grow Connection}. \normalfont In order to showcase the four architectural mutations effectively, we elaborate in the next four figures over a hypothetical scenario in which we require a neural network to input three and output two values. In such a case, a network is initialized with three input nodes and two output nodes, all devoid of connections. Calling the “Grow Connection” mutation on this network requires sampling a first node from the three input nodes and a second node from the two output nodes. Imagining these turn out to be the third input node and the second output node, a weighted connection is created in between the two.}
\label{fig:grow_connection}
\end{figure}

\end{minipage}
\hspace{0.02\linewidth}
\begin{minipage}{0.484\linewidth}

\subsection{Mutation \#2 : Prune Connection} \label{prune_connection}
\textbf{Step 1.} A first node is sampled from the list of emitting nodes.\\
\textbf{Step 2.} A second node is sampled from the first node’s set of out-nodes.\\
\textbf{Step 3.} The existing connection between the first and second node is deleted.\\\\
We give an example of this mutation in \textcolor{violet}{Figure \ref{fig:prune_connection}}.
\begin{figure}[H]
\includegraphics[width=\linewidth]{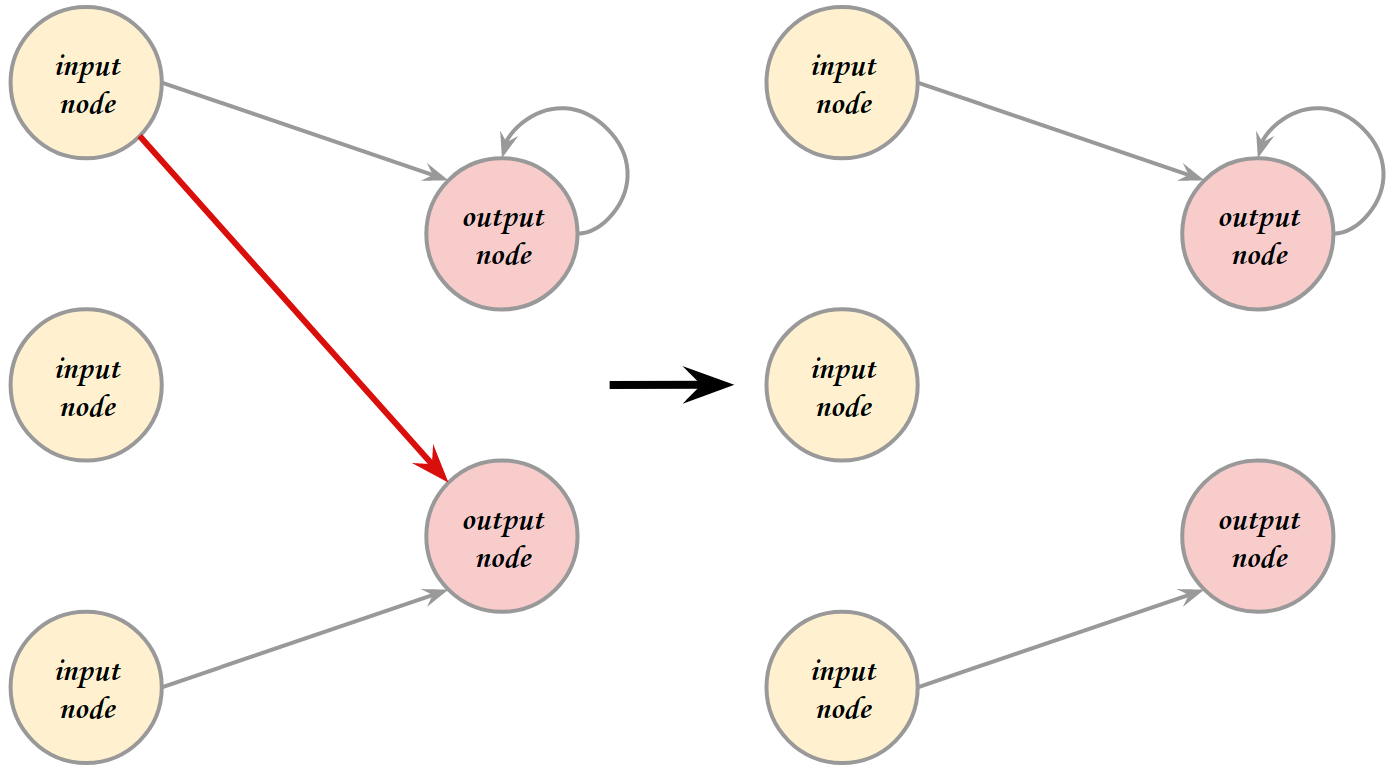}
\caption{\textbf{Mutation \#2 : Prune Connection}. We pursue the hypothetical scenario introduced in \textcolor{violet}{Figure \ref{fig:grow_connection}} and assume that three more “Grow Connection” mutations have been applied since. Calling the “Prune Connection” mutation on this network requires sampling a first node from the list composed of the first input node (2x), the third input node and the first output node. In the case that this first node turns out to be the first input node, a second node is now to be sampled from this node’s set of out-nodes, set composed of the two output nodes. We now imagine that the second output node is the one sampled. As a result, the connection between the first input node and the second output node is deleted.}
\label{fig:prune_connection}
\end{figure}

\end{minipage}
\end{minipage}

\footnotetext[2]{In this framework, sampling from sets and lists is always uniform.}
\footnotetext{Where $\mathcal{N}(\mu,\sigma^{2})$ is the Normal distribution with mean $\mu$ and variance $\sigma^{2}$, all new network weights and biases are initialized with values sampled from $\mathcal{N}(0,1)$ and perturbed every iteration with values from $\mathcal{N}(0,.01)$.}

\begin{minipage}{\linewidth}
\centering
\begin{minipage}{0.484\linewidth}

\subsection{Mutation \#3 : Grow Node} \label{grow_node}
\textbf{Step 1.} Three nodes are sampled:\\
- a first node from the set of all receiving nodes.\\
- a second node from the set of all receiving nodes minus the first node.\\
- a third node from the set of all hidden and output nodes.\\
\textbf{Step 2.} A new hidden node is initialized\footnotemark.\\
\textbf{Step 3.} Three new connections are grown:\\
- from the first node to the new hidden node.\\
- from the second node to the new hidden node.\\
- from the new hidden node to the third node.\\\\
We give an example of this mutation in \textcolor{violet}{Figure \ref{fig:grow_node}}.

\end{minipage}
\hspace{0.02\linewidth}
\begin{minipage}{0.484\linewidth}

\subsection{Mutation \#4 : Prune Node} \label{prune_node}
\textbf{Step 1.} A hidden node is sampled from the set of all hidden nodes.\\
\textbf{Step 2.} The hidden node and all of its connections are deleted.\\\\
Additionally, any hidden node, which as a result of a pruning mutation no longer receives nor emits information is also subsequently deleted.\\\\
We give an example of this mutation in \textcolor{violet}{Figure \ref{fig:prune_node}}.
\vspace{2.1em}

\end{minipage}
\end{minipage}

\footnotetext{In our implementation, the new hidden node is always positioned one layer past the first in-node, towards the out-node. If no layer exists between those two nodes, a new one is created to fit the new hidden node.}

\begin{minipage}{\linewidth}
\centering
\begin{minipage}{0.492\linewidth}

\begin{figure}[H]
\centering
\includegraphics[width=\linewidth]{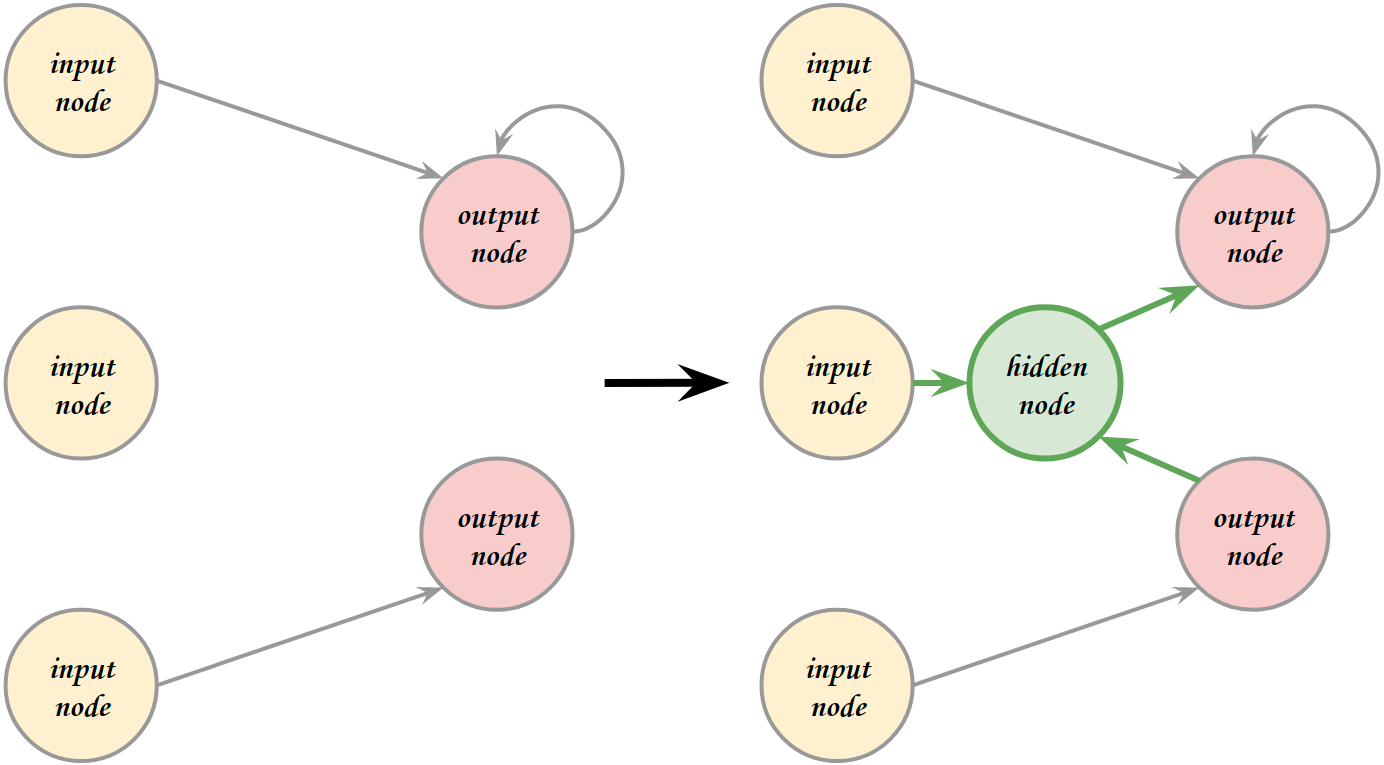}
\caption{\textbf{Mutation \#3 : Grow Node}. \normalfont We continue from where we left off in \textcolor{violet}{Figure \ref{fig:prune_connection}}. Calling the “Grow Node” mutation on this network entails sampling 1) a first node from all visible nodes 2) a second node from all visible nodes excluding that first sampled node 3) a third node from the two output nodes. We imagine these turn out to be the second input node, the second output node and the first output node respectively. A hidden node is thus created in between and connections are grown from 1) the second input node to the hidden node 2) the second output node to the hidden node 3) the hidden node to the first output node.}
\label{fig:grow_node}
\end{figure}

\end{minipage}
\hspace{0.02\linewidth}
\begin{minipage}{0.476\linewidth}

\begin{figure}[H]
\centering
\includegraphics[width=\linewidth]{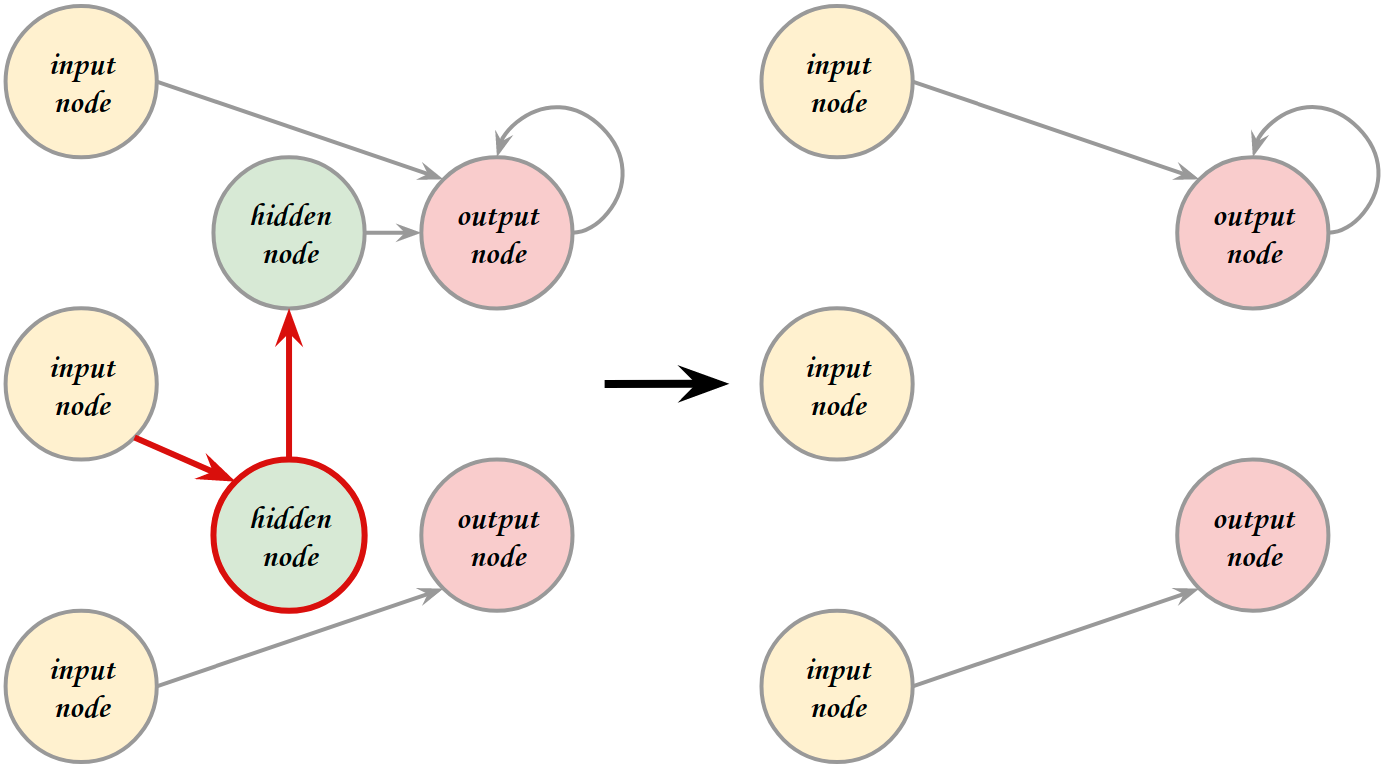}
\caption{\textbf{Mutation \#4 : Prune Node}. We now imagine that many of the three mutations described thus far have taken place since \textcolor{violet}{Figure \ref{fig:grow_node}}. Calling the “Prune Node” mutation on this network requires sampling one node from the two hidden nodes. In the case that it turns out to be the second hidden node, it is therefore deleted alongside its two connections. As a result of this deletion, the first hidden node, now devoid of any input information, is also deleted.}
\label{fig:prune_node}
\end{figure}

\vspace{2.9em}

\end{minipage}
\end{minipage}

\begin{algorithm}[H]
\caption{Neuroevolution Algorithm }\label{alg:alg}
\begin{algorithmic}
\State {\bfseries Input :} number of generations $G$, population $P$
\State - - - - - - - - - - - - - - - - - - - - - - - - - - - - - - - - - - - - - - - - - - - - - - - - - - - - - - - - - - - - - - - - - - - -
\State Initialize all agents in $P$
\For{generation = $1,2,...,G$}
\For{agent \textbf{in} $P$}
\State \textbf{\textcolor{teal}{Variation} :} Perturb the agent's weights and biases
\State \;\;\;\;\;\;\;\;\;\;\;\;\;\;\;+ Sample and apply 1 architectural mutation \textit{\footnotesize (dynamic network populations only)}
\State \textbf{\textcolor{olive}{Evaluation} :} Run the agent and track its fitness
\EndFor
\State \textbf{\textcolor{brown}{Selection} :} $P \leftarrow$ Agents with top 50\% fitness duplicated
\EndFor
\end{algorithmic}
\end{algorithm}

\section{Experiments}
\vspace{-.5em}
In order to evaluate our methods, we propose to run experiments on all 19 state-based control task environments currently available through the OpenAI Gym library \citep{brockman2016openai}. Throughout these environments, agents get to control virtual bodies like simplified robotic arms, vehicles and various types of androids. In order to perform the control tasks, agents are iteratively fed input values characterizing various pieces of information such as the position, angle and speed of their body parts. In turn, agents are expected to output values representing various actuation forces onto their body components. As a means to drive behaviour in reinforcement learning settings, the environments define termination criteria and reward signals which we both make use of in order to evaluate our agents.

To optimize the agents, we set up a simple neuroevolution algorithm as described in \textcolor{violet}{Algorithm \ref{alg:alg}}. In order to better understand the value brought forward by our framework, we propose to not only evolve dynamic networks but also standard deep recurrent neural networks of dimensions \mbox{$[d\_input, 50, 50, d\_output]$}, similar to the networks described by \cite{salimans2017evolution} for their MuJoCo experiments.

After choosing a population size and a task to optimize for, the evolutionary optimization process begins by initializing a population of network agents. While no architectural operation is required in static-sized deep neural networks, the dynamic networks are always set to first create the minimal amount of structure to fit the task at hand, as described in \textcolor{violet}{Figure \ref{fig:grow_connection}}. Then, regardless of dynamicity, all network weights and biases are initialized to zero. The evolutionary process then starts iterating over three distinct stages commonly referred to as variation, evaluation and selection.

First, during the variation stage, we begin by randomly perturbing the parameters of both dynamic and static networks. Like for dynamic networks, we propose to perturb static networks with values drawn from the normal distribution $\mathcal{N}(0, 0.01)$. As described in \textcolor{violet}{Section \ref{methods}}, dynamic agents then sample one of four architectural mutations, which they in turn apply to their existing network.

Then, during the evaluation stage, agents run one episode\footnote{Except for environments \textit{Pendulum} and \textit{Reacher}, in which we run five episodes to more accurately evaluate overall agent performance given their large task initialization variance.} in the environment until termination. In environments requiring continuous action values, we clip the ReLU activated values emitted from the networks' output nodes to the range [0, 1] and scale them to the expected range of outputs. In environments requiring discrete action values, we instead retrieve the position of the output node emitting the largest value. Lastly, while we directly feed agents with values emitted from the environment in most tasks, early experiments found the distribution of input values to hinder progress in certain environments. We therefore put in place a running standardization of inputs for classical task \textit{Pendulum} and MuJoCo tasks \textit{Ant}, \textit{HalfCheetah}, \textit{Hopper}, \textit{Humanoid}, \textit{Reacher} and \textit{Walker2d}.

Finally, during the selection stage, we utilize a 50\% truncation strategy, in which agents with upper half performance in terms of accumulated reward are maintained and duplicated. Agents then proceed back to the variation stage and keep iterating until termination. We fully parallelize the variation and evaluation stages of our algorithm and run the entirety of these experiments on a cluster with nodes powered by AMD 7532 CPUs, connected through NVIDIA QM8700 switches. We provide information about computation speed in \textcolor{violet}{Table \ref{table:table}} plus the source code, final model checkpoints and results at \textcolor{blue}{\href{http://www.github.com/MaximilienLC/nra/}{github.com/MaximilienLC/nra/}}.

\vspace{-.5em}
\section{Results}
\vspace{-.5em}

We display in \textcolor{violet}{Figures \ref{fig:classic}} to \textcolor{violet}{\ref{fig:mujoco_2}} the evolution of scores achieved by various static and dynamic network populations on all 19 tasks. These scores were obtained by averaging performance over 10 newly seeded runs\footnote{We iterate over seeds starting from 0 during the evolutionary optimization and make use of seeds $2^{31}$-10 to $2^{31}$-1 at test time.}. We complement these figures with a representation of the final dynamic elite (highest scoring) architecture for each task.

We first observe that across all tasks, neural network agents with dynamic architectures always match and often even outperform (ex: \textit{BipedalWalker} and \textit{Reacher}) evolved static deep neural network agents. Moreoever, we find cases in which dynamic agents require smaller population sizes (ex: \textit{MountainCar}), a lower amount of iterations (ex: \textit{Acrobot} and \textit{CartPole}) or even both (ex: \textit{InvertedDoublePendulum} and \textit{InvertedPendulum}) in order to reach what we believe to be optimal performance. We also find cases where static network agents, as opposed to dynamic agents, no longer appear to be capable of making sustained progress (ex: \textit{Ant} and \textit{HalfCheetah}).

Secondly, and perhaps more importantly, we find that out of the 19 control tasks, the dynamic network agents either match or surpass the majority of the baselines on 13 tasks. Out of the remaining 6 tasks, dynamic agents still appear to be trending upwards (at varying speed) after 5000 generations on 5 tasks but seem unable to make further progress on the task \textit{BipedalWalkerHardcore}.

Finally, we find from the third column of these figures that the number of parameters utilized by elite dynamic agents is extremely condensed by modern standards, ranging from 3 parameters on \textit{MountainCarContinuous} to 322 parameters on \textit{Pendulum}, averaging at about 127 parameters across all tasks. We notice quite different architectures evolved across different tasks, some not requiring hidden nodes (ex: \textit{CartPole} and \textit{MountainCarContinuous}), some that are much deeper than they are wide (ex: \textit{HalfCheetah} and \textit{Reacher}) and others with early layers much wider than later ones (ex: \textit{Ant} and \textit{Humanoid}). Generally though, network layers tend to be very interconnected through both forward and recurrent connections.

\begin{minipage}{\linewidth}
\centering
\begin{minipage}{0.484\linewidth}

\begin{figure}[H]
  \centering
  \includegraphics[width=\linewidth]{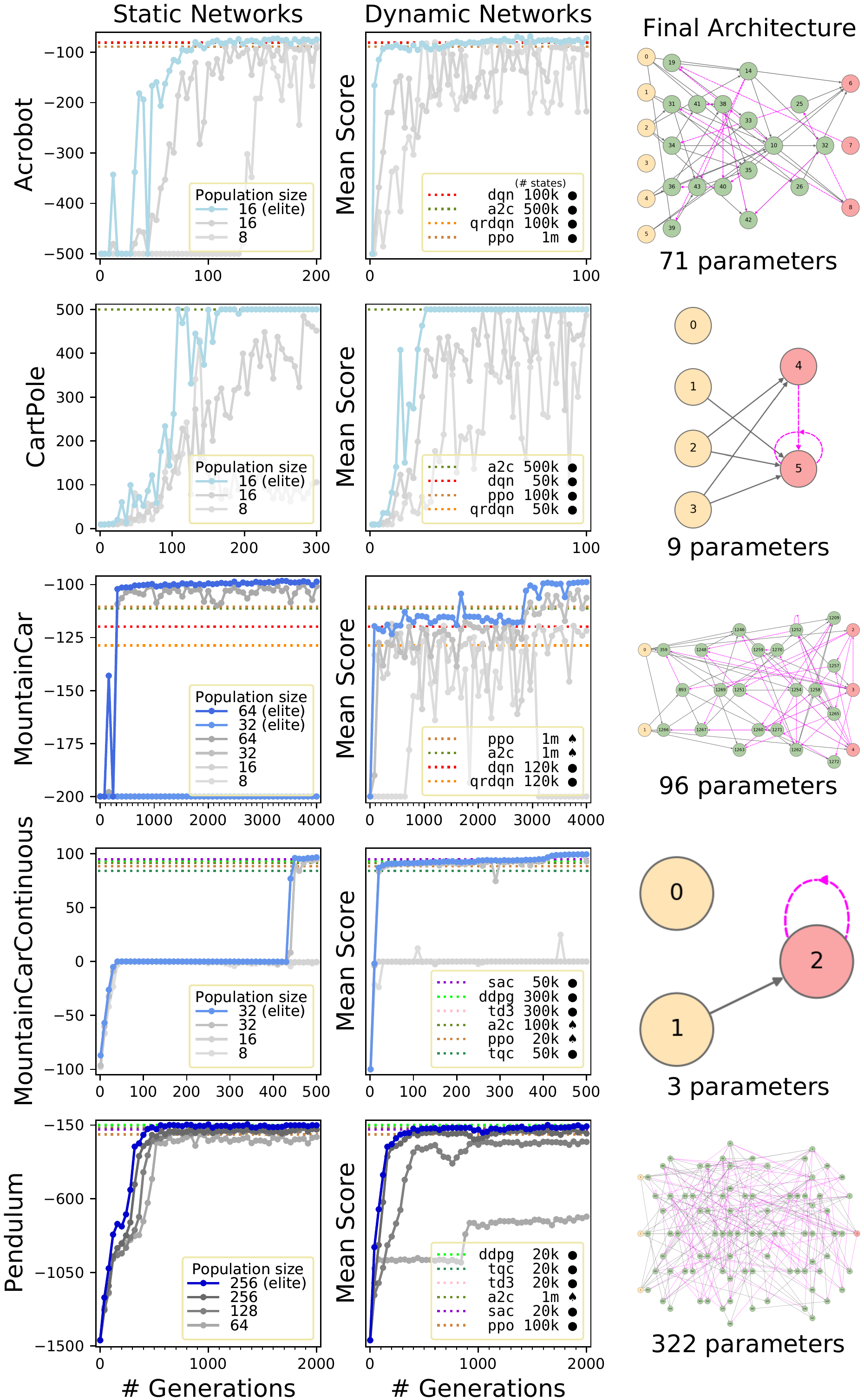}
  \caption{\textbf{Classic Control Tasks Results}. \normalfont (Baseline results marked with \ding{108} are reported from evaluating pre-trained agents \cite{rl-zoo} on the same seeds as our agents. Baseline results marked with \ding{171} are directly reported from \cite{rl-zoo}). Both static and dynamic networks match or surpass the baselines on classic control tasks.}
  \label{fig:classic}
\end{figure}

\end{minipage}
\hspace{0.02\linewidth}
\begin{minipage}{0.484\linewidth}
\vspace{2.65em}
\begin{figure}[H]
  \centering
  \includegraphics[width=\linewidth]{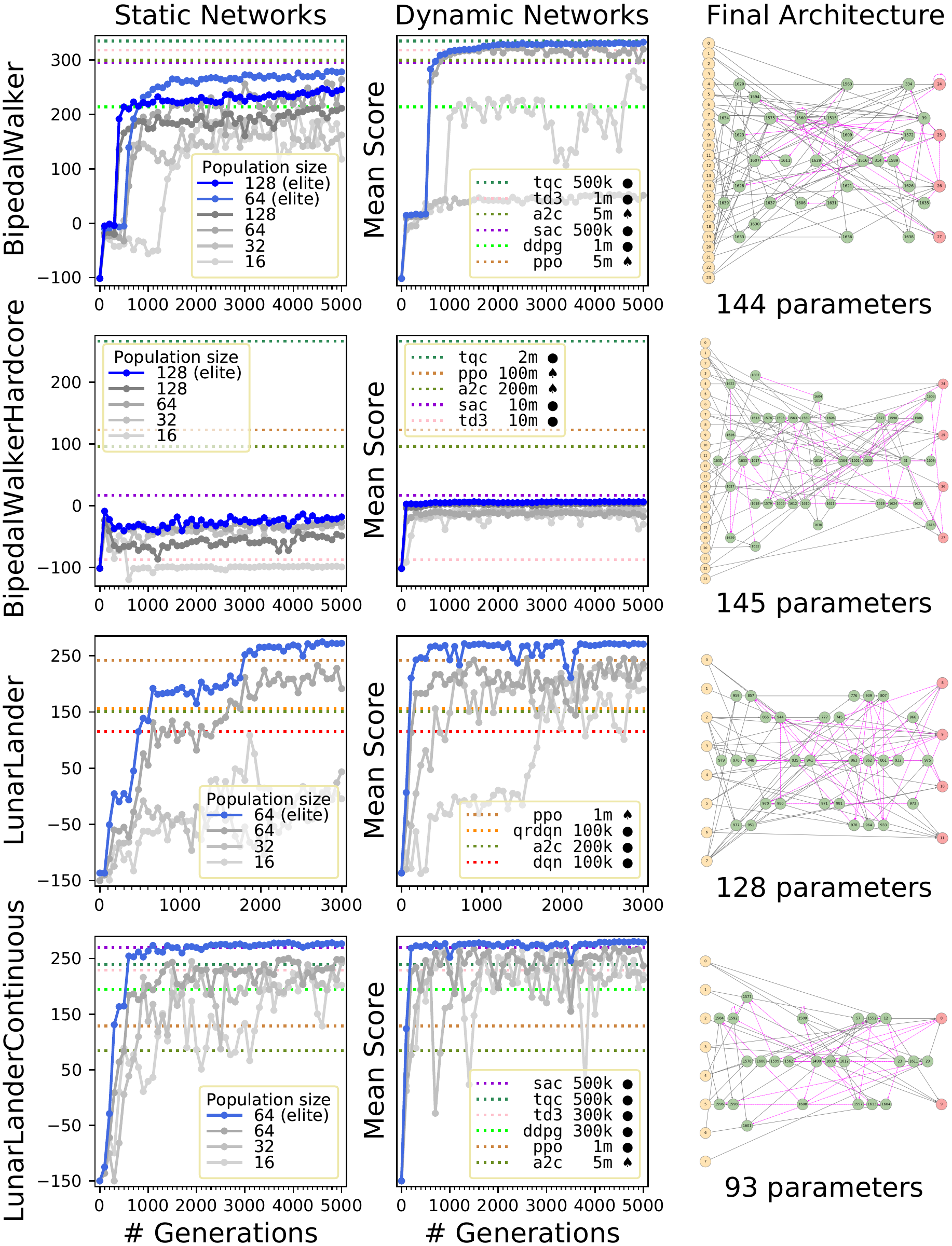}
  \caption{\textbf{Box2D Control Tasks Results}. \normalfont Dynamic networks perform as well or better than static networks on all Box2D tasks. They are also competitive with the baselines on all tasks except \textit{BipedalWalkerHardcore}.}
  \label{fig:box2d}
\end{figure}

\end{minipage}
\end{minipage}

\begin{minipage}{\linewidth}
\centering
\begin{minipage}{0.484\linewidth}

\begin{figure}[H]
  \centering
  \includegraphics[width=\linewidth]{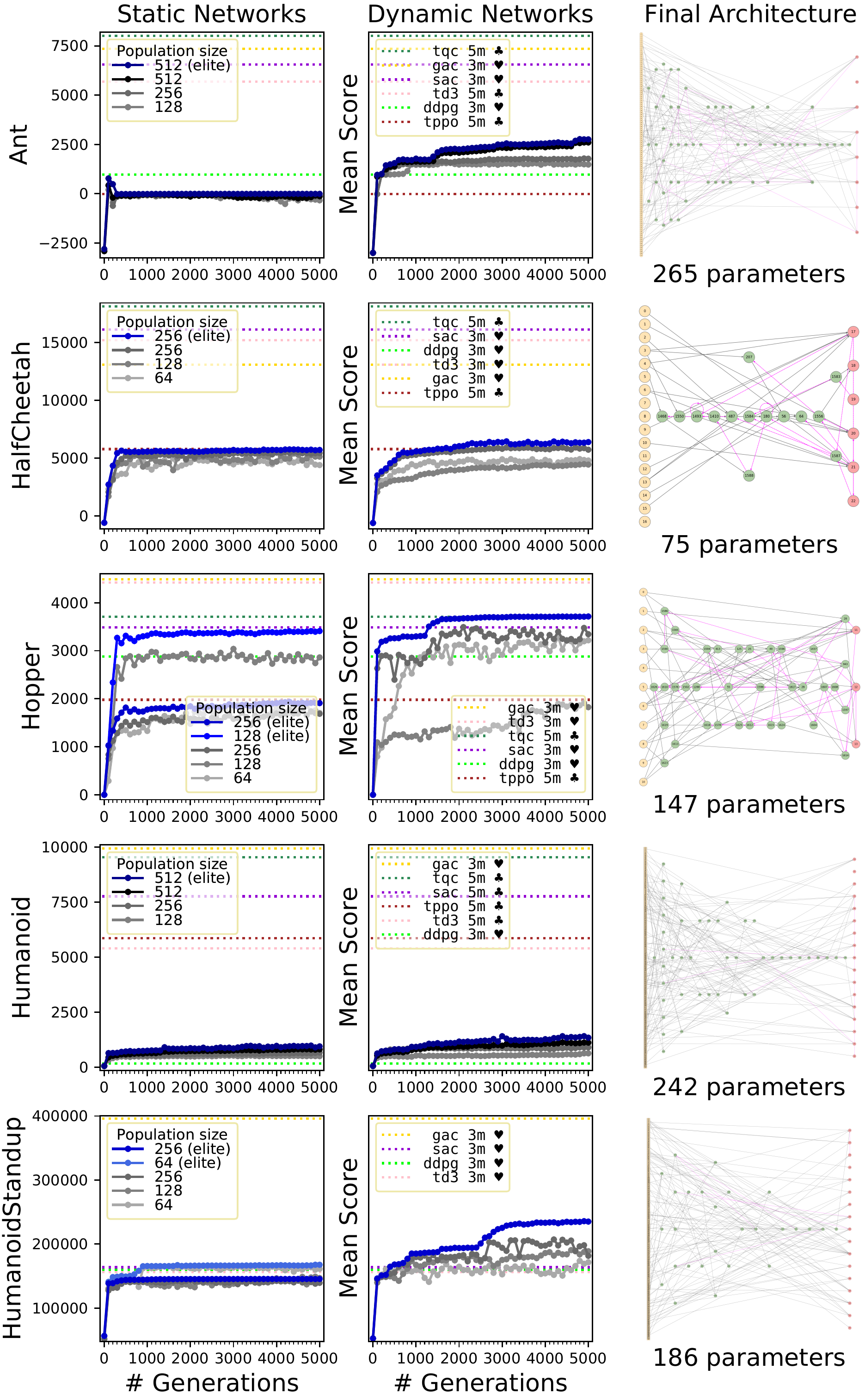}
  \caption{\textbf{MuJoCo Control Tasks Results (1/2)}. \normalfont (Baselines marked with \ding{168} and \ding{170} are reported from \cite{kuznetsov2020controlling} and \cite{lingwei2021generative} respectively) Dynamic networks perform better than static networks across this first batch of MuJoCo tasks. Performance matches or exceeds most of the baselines on \textit{Hopper} and \textit{HumanoidStandup}. Progress is slower on \textit{Ant}, \textit{Half-Cheetah} and \textit{Humanoid}, not quite reaching the same level of performance in 5000 generations.}
  \label{fig:mujoco_1}
\end{figure}

\end{minipage}
\hspace{0.02\linewidth}
\begin{minipage}{0.484\linewidth}

\begin{figure}[H]
  \centering
  \includegraphics[width=\linewidth]{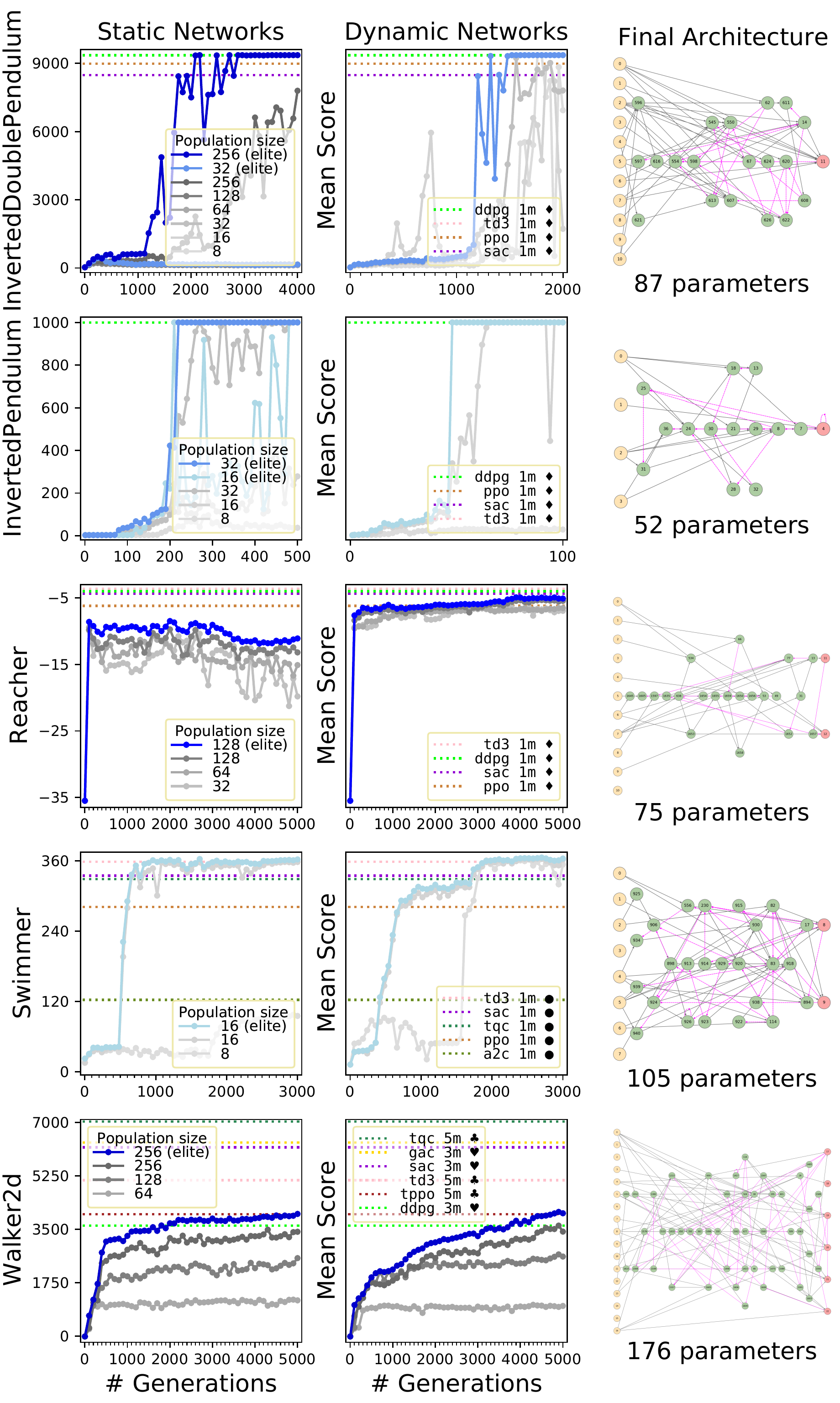}
  \caption{\textbf{MuJoCo Control Tasks Results (2/2)}. \normalfont (Baselines marked with \ding{169} are reported from \cite{fujimoto2018addressing}) Dynamic agents perform as well or better than static agents on the remaining MuJoCo tasks. Performance comes close to matching, matches or exceeds the baselines on all tasks except \textit{Walker2d} where it has yet to surpass the majority of theses techniques after 5000 generations.}
  \label{fig:mujoco_2}
\end{figure}

\vspace{1.9em}

\end{minipage}
\end{minipage}

\vspace{-2em}
\section{Discussion}
\vspace{-.5em}

We have presented in this work recurrent neural networks whose architectures dynamically evolve in capacity during the optimization process. We made use of a simple yet scalable evolutionary algorithm and experimented on many continuous and discrete control tasks. We first found this framework to evolve networks that are generally more capable than evolved static-sized networks, leading us to believe that starting off with the smallest possible architecture and progressively constructing it enables agents to better adapt to each of these tasks. In addition, we found the dynamic networks to have very interconnected layers and various types of recurrences which appear to grant networks more representational flexibility and thus have them require utilizing fewer parameters. Since we did not perform any hyperparameter search, it is quite possible that static networks would perform better with different hyperparameters on many of these tasks. However it seems equally likely that dynamic networks would also benefit from task specific hyperparameters such as sampling more mutations per iteration in complex \mbox{MuJoCo} tasks. The focus of this work was rather to explore whether a non-specialized approach could perform well across many tasks.

We then found the dynamic architecture framework (and the static framework to a lesser degree) to produce network agents that are most often competitive with the deep reinforcement learning agent baselines. And while progression remains quite slow for several tasks, not quite reaching most baselines’ performance after 5000 generations, their progression asymptote often suggests that the evolutionary optimization has yet to converge. In addition to longer run times, results from \cite{such2017deep} that surpass the majority of the baselines presented in this work on the task \textit{Humanoid} seem to suggest that larger population sizes and more specialized hyperparameters could provide further improvements. Finally, we remark that modern reinforcement learning techniques are often evaluated in terms of data efficiency and therefore cannot guarantee that these baselines correspond to some of these agents' optimal performance. Our technique often requires running a much larger number of environment steps and it could be argued that deep reinforcement learning algorithms would make better use of running these many states. However, evolutionary algorithms such as the one used in this work are by design very straightforward to massively parallelize and can therefore attain this amount of states in reasonable run times without even requiring the use of GPUs.

Importantly however, as \cite{such2017deep} discovered in their Atari experiments, we encounter tasks (e.g. \textit{BipedalWalkerHardcore}) in which agents are noticeably unable to make sustained progress over thousands of generations, which further suggests that as a drawback of its simplicity, this evolutionary optimization formulation is considerably affected by unfavorable reward landscapes that are too coarse to drive the desired change in behaviour. Exploring other evolutionary learning paradigms to produce behaviour in such settings therefore seems like a worthwhile future direction.

\vspace{-.5em}
\section{Conclusion}
\vspace{-.5em}

In addition to joining earlier works in suggesting that evolutionary algorithms are a competitive alternative for optimizing neural network parameters, we believe our work to be a new step forward in demonstrating the feasibility of evolving network architectures in complex machine learning settings. We find clear advantages to constructing architectures like requiring orders of magnitude fewer parameters in order to reach very high performance on many reinforcement learning state-based control tasks. And while our results do not achieve state-of-the-art performance on every task that we experiment on, we believe the conceptual simplicity of our method to give room for many types of improvements. Moving forward, leveraging the continually expanding representation space of such dynamic networks could prove to be valuable in order to explore both continual and multi-task learning settings. In the meantime, we believe that our framework could prove useful in real-life applications where network compactness and autonomous design are of critical importance.

\renewcommand{\arraystretch}{.9}
\begin{table}[H]
  \centering
  \caption{\textbf{Runtime, number of visited states and elite behaviour of the largest dynamic network populations on each task.}}
  \label{tab:freq}
  \setlength\tabcolsep{2.1pt}
  \begin{tabular}{|c|ccc|}
    \hline
    Task & Runtime (cores) & States & Behaviour\\
    \hline
    Acrobot-v0 & 0h01m \hspace{0.15cm}(16) & $\sim$138K & [\textcolor{blue}{\underline{\href{https://drive.google.com/file/d/1thyWIyHvT4j_F9owgBghHrQhJPA1myJJ/view}{video}}}]\\
    CartPole-v1 & 0h02m \hspace{0.15cm}(16) & $\sim$657K & [\textcolor{blue}{\underline{\href{https://drive.google.com/file/d/1F-pJFmrcKw96c-uDSIXkX4oXrUOc1S3y/view}{video}}}]\\
    MountainCar-v0  & 0h06m \hspace{0.15cm}(32) & $\sim$15M & [\textcolor{blue}{\underline{\href{https://drive.google.com/file/d/1vmOv5tIlezcDJca5Zh7rNbdxpRShgvj0/view}{video}}}]\\
    MountainCarContinuous-v0 &  0h02m \hspace{0.15cm}(32) & $\sim$4M & [\textcolor{blue}{\underline{\href{https://drive.google.com/file/d/1JURqnZ8I2iwCFnd6Nc5XJS7o4-1ucalp/view}{video}}}]\\
    Pendulum-v1 & 2h13m (256) & $\sim$512M & [\textcolor{blue}{\underline{\href{https://drive.google.com/file/d/1e_34-BfUhQgm0FHRMSzg66Gjx1koL4tr/view}{video}}}]\\
    BipedalWalker-v3 & 3h01m \hspace{0.15cm}(64) & $\sim$419M & [\textcolor{blue}{\underline{\href{https://drive.google.com/file/d/1LW6MEzKzoSu3z1pZR0QtD-QalRWoIoyV/view}{video}}}]\\
    BipedalWalkerHardcore-v3 & 3h28m (128) & $\sim$1B & [\textcolor{blue}{\underline{\href{https://drive.google.com/file/d/1th6M8PEjXgaIC1Lq_eR8symzcRe2wL9v/view}{video}}}]\\
    LunarLander-v2 & 1h15m \hspace{0.15cm}(64) & $\sim$48M & [\textcolor{blue}{\underline{\href{https://drive.google.com/file/d/14keCx-ZZ1vJ3J_oQ3JfK0sMWtZClfFuh/view}{video}}}]\\
    LunarLanderContinuous-v2 & 3h16m \hspace{0.15cm}(64) & $\sim$62M & [\textcolor{blue}{\underline{\href{https://drive.google.com/file/d/1xT58dhP8n_gr0kemk2f6gY4BVspClEx6/view}{video}}}]\\
    Ant-v3 & 2h49m (512) & $\sim$3B & [\textcolor{blue}{\underline{\href{https://drive.google.com/file/d/1S6bF6xlUvQ78X6rizJfYdKcjlwHIDU2z/view}{video}}}]\\
    HalfCheetah-v3 & 1h56m (256) & $\sim$1B & [\textcolor{blue}{\underline{\href{https://drive.google.com/file/d/1IqYasTdzYX5iAdwRXFDY1yO_CNM-Rihk/view}{video}}}]\\
    Hopper-v3 & 2h35m (256) & $\sim$1B & [\textcolor{blue}{\underline{\href{https://drive.google.com/file/d/1WYOF5j2UZS4qhNo_rpqXwsbY4cq8S2MA/view}{video}}}]\\
    Humanoid-v3 & 2h00m (512) & $\sim$532M & [\textcolor{blue}{\underline{\href{https://drive.google.com/file/d/192NcLvpOLiXzqHOhJdaxsnGgeiPa_X6N/view}{video}}}]\\
    HumanoidStandup-v2 & 7h04m (256) & $\sim$1B & [\textcolor{blue}{\underline{\href{https://drive.google.com/file/d/1dsHZ3SveQiNHbHeyu4Jut25jTZuowM8a/view}{video}}}]\\
    InvertedDoublePendulum-v2 & 0h09m \hspace{0.15cm}(32) & $\sim$21M & [\textcolor{blue}{\underline{\href{https://drive.google.com/file/d/1AsGQq2kQq1_mlDhqmDo2k2zJXL0jkstN/view}{video}}}]\\
    InvertedPendulum-v2 & 0h01m \hspace{0.15cm}(16) & $\sim$872K & [\textcolor{blue}{\underline{\href{https://drive.google.com/file/d/15XMvygLxbhTnqGgu9ZuGdj5yx1DXQbF5/view}{video}}}]\\
    Reacher-v2 & 0h31m (128) & $\sim$160M & [\textcolor{blue}{\underline{\href{https://drive.google.com/file/d/1kzQHMLarPYZU2Whgqg8clhXe3G4agHa1/view}{video}}}]\\
    Swimmer-v3 & 0h49m \hspace{0.15cm}(16) & $\sim$48M & [\textcolor{blue}{\underline{\href{https://drive.google.com/file/d/1GEUNlRKI6f7YR_90GGh9qjY6-z4ZHuWf/view}{video}}}]\\
    Walker2d-v3 & 3h00m (256) & $\sim$1B & [\textcolor{blue}{\underline{\href{https://drive.google.com/file/d/1PjapO6V7l_0_rF4LI0FWKiTZd-jLmrBO/view}{video}}}]\\
  \hline
\end{tabular}
\label{table:table}
\end{table}

\bibliography{iclr2022_conference}
\bibliographystyle{iclr2022_conference}

\end{document}

%% file: math_commands.tex
\usepackage{amsmath,amsfonts,bm}

\def\eqref#1{equation~\ref{#1}}
\def\1{\bm{1}}

\DeclareMathAlphabet{\mathsfit}{\encodingdefault}{\sfdefault}{m}{sl}
\SetMathAlphabet{\mathsfit}{bold}{\encodingdefault}{\sfdefault}{bx}{n}